\documentclass{article}

\usepackage{arxiv}
\usepackage{kotex}
\usepackage[utf8]{inputenc} 
\usepackage{multirow}
\usepackage[T1]{fontenc}    
\usepackage{hyperref}       
\usepackage{url}            
\usepackage{booktabs}       
\usepackage{amsfonts}       
\usepackage{nicefrac}       
\usepackage{microtype}      
\usepackage{lipsum}		
\usepackage{graphicx}
\usepackage{natbib}
\usepackage{doi}
\title{Deep Neural Networks and End-to-End Learning for Audio Compression}

\date{} 					

\author{ Daniela N. Rim \\ 
	\textit{Handong Global University}\\
	Republic of Korea\\
	\And
	Inseon Jang \\
   \textit{ Electronics and Telecommunications } \\\textit{Research Institute }\\
    Republic of Korea\\
	\And
    Heeyoul Choi \\
	\textit{Handong Global University}\\
	Republic of Korea\\
	\texttt{heeyoul@gmail.com} \\
}

\renewcommand{\headeright}{}
\renewcommand{\undertitle}{}

\hypersetup{
pdftitle={A template for the arxiv style},
pdfsubject={cs.ML},
pdfauthor={Daniela Rim},
pdfkeywords={Audio Compression, End-to-End Learning, Discrete Latent Space, Variational Autoencoder},
}

\begin{document}
\maketitle

\begin{abstract}
Recent advances in end-to-end deep learning have encouraged the exploration of tasks dealing with highly structured data using unified deep network models. The fabrication and design of such models for compressing audio signals has been a challenge due to the need for discrete representations that are not easy to train with end-to-end backpropagation. In this paper, we present an end-to-end deep learning approach that combines recurrent neural networks (RNNs) within the training strategy of variational autoencoders (VAEs) with a binary representation of the latent space. We apply a reparametrization trick for the Bernoulli distribution for the discrete representations, which allows smooth backpropagation. In addition, our approach enables the separation of the encoder and decoder, which is necessary for compression tasks. To the best of our knowledge, this is the first end-to-end learning for a single audio compression model with RNNs, and our model achieves a Signal to Distortion Ratio (SDR) of 20.53dB.
\end{abstract}


\vspace{4em}

\section{Introduction}
\par The recent improvements of deep neural network (DNN) models and learning methods have open new horizons for highly structured sequential data-related tasks, including automatic speech recognition (ASR) (\cite{hinton2012deep}) as well as audio compression (\cite{kankanahalli2018end}, \cite{zhen2019cascaded}, \cite{valin2019lpcnet}). There have been particular interests in developing end-to-end DNN models that can handle raw input audio signals with little-to-none biased intervention. For audio compression, for example,  \cite{kankanahalli2018end} proposed a complete single DNN pipeline consisting mainly of residual networks and autoencoders. Several variants of this network aim to reduce its computational cost, e.g., CMRL (\cite{zhen2019cascaded}) or LPCNet (\cite{valin2019lpcnet}).
\par The audio compression task is particularly challenging due to several issues related to the type of data and the required complexity of the model’s architecture. 
\par First, we need robust and flexible models to capture the complex relations between the data variables through time. Essentially, an end-to-end compression architecture aims to map an audio signal into a more concise portable representation. This characteristic is shared with deep generative models, targeting capturing critical aspects of data to generate new data instances (\cite{ishizone2020ensemble}, \cite{nistal2021comparing}, \cite{dhariwal2020jukebox}, \cite{dieleman2018challenge}). A notable example of this approach consists of a combination of Variational Autoencoders (VAEs) (\cite{kingma2013auto}) and Recurrent Neural Networks (RNNs) (\cite{rumelhart1986learning}, \cite{hahn2018application}). RNNs can handle extensive time-dependent data samples, but their deterministic hidden variables cannot capture the variability present in raw audio. On the other hand, VAEs provide high-level latent random variables and flexible mapping to the model’s output. It has been shown that applying a VAE to the input sequence and using an RNN to model the output distribution succeeded in the modeling and recognition of speech signals (\cite{bayer2014learning}, \cite{chung2015recurrent}, \cite{fraccaro2016sequential}, \cite{goyal2017z}). However, such method has not been studied in detail for audio compression tasks.
\par Second, when raw audio signals are encoded into discrete representation, it makes training by backpropagation unfeasible because discrete representation blocks the flow of gradient information. Some compression models circumvent this issue using vector quantization (VQ) in the coding layer (\cite{kankanahalli2018end}), which does not train the discrete representation by backpropagation.
\par Third, for real-world implementations, the encoder and decoder of the compressor should be decoupled and placed in different devices. This small detail should be taken into account when designing the architecture.
\par In this work, we propose a Variational Recurrent Encoder-Decoder (VRED), an end-to-end DNN for audio compression. In addition, we can use trainable features with convolutional and deconvolutional layers before and after VRED, respectively. Thus, the proposed network consists of a convolution layer to extract features, VRED, and a deconvolution layer to reconstruct audio signals. To make training more efficient, we train the audio compressor in three stages: (1) we pre-train the feature extractor (convolutional and deconvolutional layers), (2) we freeze these layers and train VRED, and (3) we fine-tune the whole architecture together.
\par The model presents the following benefits:
\begin{itemize}
    \item Simple objective function compared to other end-to-end models (\cite{kankanahalli2018end}).
    \item Possible for compression compared to other VAE+RNN models like Variational Recurrent Neural Networks (VRNN) (\cite{chung2015recurrent}). 
    \item More natural architecture for the discrete variables compared to VQ-VAE (\cite{oord2017neural}).
\end{itemize} 
\par In experiment results, the proposed network achieves a signal-to-distortion ratio (SDR) of 20.53, with a compression ratio of 1/11. To our knowledge, this is the first approach that uses an entirely separable audio compression model using an end-to-end VAE-RNN model.

\section{Preliminaries}
\label{sec:prel}

\subsection{Recurrent Neural Networks}
Given a sequential input $\mathbf{x}$ of length $T$, a simple RNN recursively updates its intermediate hidden state layer by
\begin{equation}\label{eq11}
    h_t = f_\theta(x_t, h_{t-1}),
\end{equation}
for each $x_t \in \mathbf{x}$ using a $\theta-$parameterized function $f$. This deterministic non-linear transition function $f$ can be implemented using gated mechanisms such as GRUs, LSTMs, and so on (\cite{cho2014learning}, \cite{hochreiter1997long}, \cite{choi2019persistent}). 
\par RNNs model the sequences of probability distribution by the joint distribution:
\begin{equation} \label{eq2}
    p(x_1,x_2,...,x_T) = \prod_{t=1}^{T} p(x_t|x_{<t}). 
\end{equation}
\par Each conditional probability $p(x_t|x_{<t})$ in Eq. \ref{eq2} is represented by the output of an RNN at a single step. We can interpret them as a parameterized function $d_{\tau}$ that maps the RNN hidden state $h_{t-1}$ to a probability distribution over possible outputs: $p(x_t|x_{<t}) = d_{\tau}(h_{t-1})$. This function $d_{\tau}$ is responsible for the representational power of an RNN, as it contains information of previous sequences.  

\subsection{Variational Autoencoder}
\par VAEs (\cite{kingma2013auto}) are responsible for the expressivity in the model’s latent space, as they aim to capture the variations in the input variables $\mathbf{x}$ through a set of random variables $\mathbf{z}$ which aim to be disentangled (\cite{higgins2016beta}, \cite{hahn2018application}). Their goal is to train the generative model following the joint distribution $p(\mathbf{x},\mathbf{z}) = p(\mathbf{x}|\mathbf{z})p(\mathbf{z})$. Here $p(\mathbf{x}|\mathbf{z})$ is the likelihood (decoder) that generates $\mathbf{x}$ data from $\mathbf{z}$, and $p(\mathbf{z})$ is a prior over latent variables $\mathbf{z}$. 
\par For inference, input is encoded to a distribution $p(\mathbf{z}|\mathbf{x}) = p(\mathbf{x}|\mathbf{z}) p(\mathbf{z}) / p(\mathbf{x})$. The true posterior $p(\mathbf{z}|\mathbf{x})$ is intractable, so the model parameterizes an approximate posterior distribution (encoder) $q(\mathbf{z}|\mathbf{x})$ instead. In most applications, the parameters of the encoder, decoder, and prior are computed using neural networks. Usually, the prior is chosen to be normally distributed with diagonal covariance, which allows for the Gaussian reparameterization trick to be used to compute the gradient of the objective function (\cite{kingma2013auto}, \cite{higgins2016beta}). 
\par The use of the approximate posterior $q(\mathbf{z}|\mathbf{x})$ allows VAEs the use of the Evidence Lower Bound (ELBO):
\begin{equation} \label{elbo}
    \log p(x) \geq -KL(q(\mathbf{z}|\mathbf{x})||p(\mathbf{x})) + \mathbb{E}_{q(\mathbf{z}|\mathbf{x})} [\log p(\mathbf{x}|\mathbf{z})],
\end{equation}
where $KL$ is the Kullback-Leibler divergence between two distributions. 

\subsection{Variational RNN}
\begin{figure}[ht]
	\centering
	\includegraphics[width=0.50\linewidth]{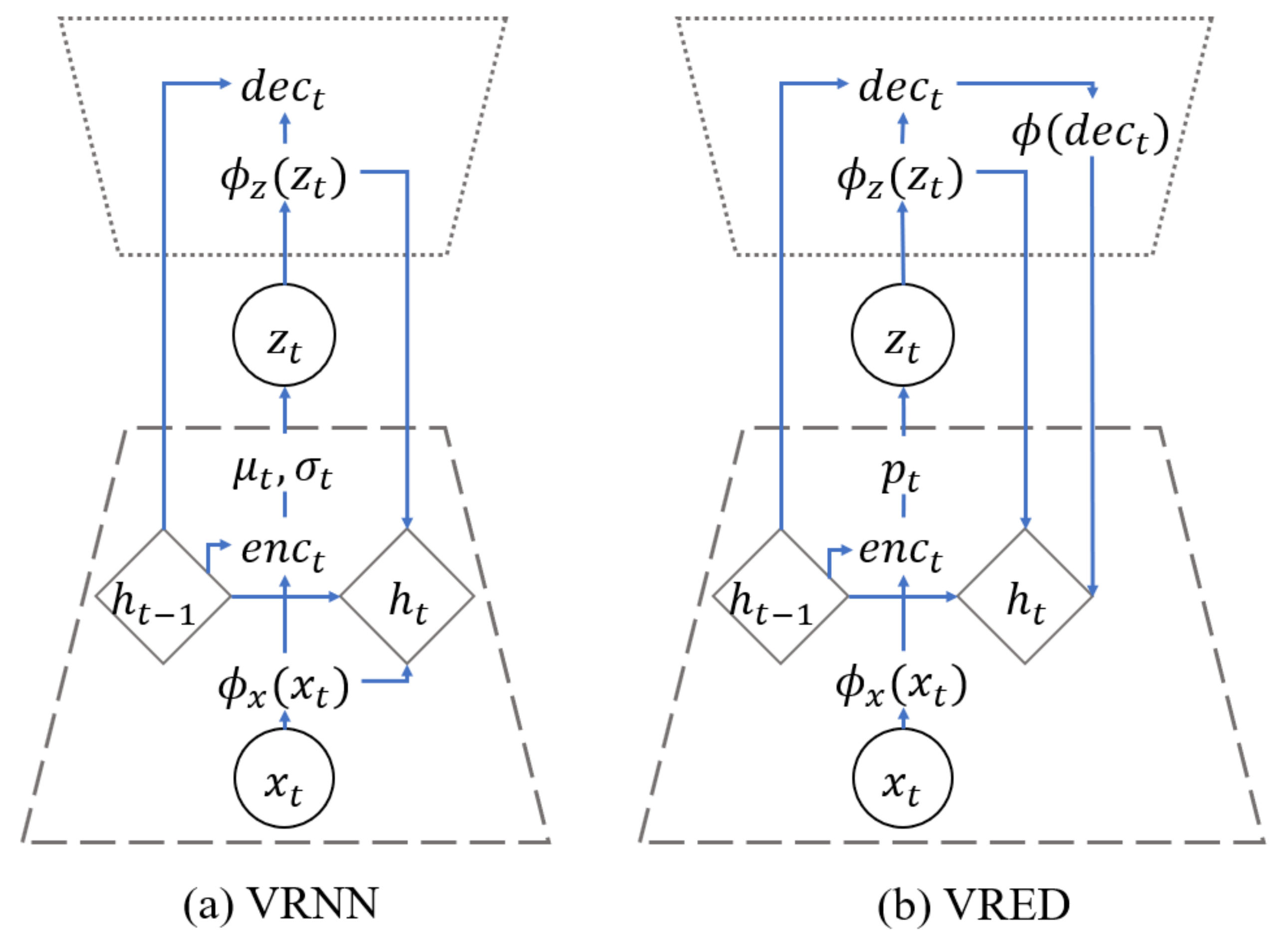}
	\caption{Difference between the training of VRNN (a) and our VRED model (b). The dashed square represents the encoder and the pointed square, the decoder.}
	\label{fig:fig1}
\end{figure}

Out of all the previous VAE-RNN models, VRNN (\cite{chung2015recurrent}) has the most similar structure to ours. The prior on the latent variable $z_t$ used in VRNN follows a normal distribution parametrized by a function of the previous hidden state $h_{t-1}$ of the RNN. During the generation, the decoder is influenced by both $h_{t-1}$ and $z_t$. Similarly, during inference, the approximate posterior depends on the input $x_t$ and on $h_{t-1}$. The general structure of the model is shown in Figure \ref{fig:fig1} (a). 
The novelty this model introduces is that the prior distribution of the latent random variable at timestep $t$ is dependent on all the preceding inputs via the RNN hidden state $h_{t-1}$. Thus, the prior distribution also includes temporal dependency. 
\par If we consider $\mathbf{z}$ as a sequence $\mathbf{z}=(z_1,z_2,...,z_T)$, the ELBO of Eq. \ref{elbo} becomes: 
\begin{equation} \label{elbo2}
 \mathbb{E}_{q(\mathbf{z}_{\leq T}|\mathbf{x}_{\leq T})} \bigg[ \sum_{t=1}^{T}(-KL(q(\mathbf{z}_t|\mathbf{x}_{\leq t}, \mathbf{z}_{<t})||p(\mathbf{z}_t|\mathbf{x}_{<t}, \mathbf{z}_{<t})) + \log p(\mathbf{x}_t|\mathbf{z}_{\leq t}, \mathbf{x}_{<t}))\bigg].
\end{equation}
\par The training of this architecture is in core the same as the VAE. 

\subsection{VQ-VAE}
\par The main difference between our VRED model and the previous works is discrete (instead of continuous) latent representations in the model. For audio compression, discrete representations are better suited but challenging to train the model with. Vector Quantised Variational Autoencoder (VQ-VAE) (\cite{oord2017neural}) was introduced to address this issue. The model successfully includes discretization in VAEs using VQ (vector quantization), reaching a similar performance to the continuous counterparts. To backpropagate through the discrete variables, however, the model gives up the benefit of the sampling method of VAEs. In addition, the model has to learn a suitable uniform categorical prior. Due to these, the training objective changes and is no longer the ELBO.

\section{Variational Recurrent Encoder-Decoder}
Even though VRNN with RNN encoder/decoder seems to be a good architecture for audio compression, it presents two out of three issues mentioned above. It does not have a discrete latent variable with a feasible backpropagation, and it has a non-separable encoder and decoder during inference. Separating the encoding and decoding of the inputs may not seem necessary for the model to generate sequences, but in real-world applications of the audio compression task, the encoder and decoder would typically be in different devices.  
\par We propose a novel model, Variational Recurrent Encoder-Decoder (VRED) for audio compression. We redesign its encoder and decoder completely separable. Furthermore, we propose the use of discrete latent variables based on the Bernoulli distribution, with a reparameterization trick that allows a smooth backpropagation. The variational inference on VRED is the same as on VAE (\cite{kingma2013auto}.)
\par Since we have feature extraction and reconstruction layers in addition to VRED, for more efficient training, the learning process can be decomposed into three stages. First, we pre-train signal extractor-constructor layers to learn the most significant features of the raw audio. Then, we freeze this feature codec and train the VRED model with the resulting feature samples. Finally, we fine-tune the whole model.

\begin{figure}[ht]
	\centering
	\includegraphics[width=0.30\linewidth]{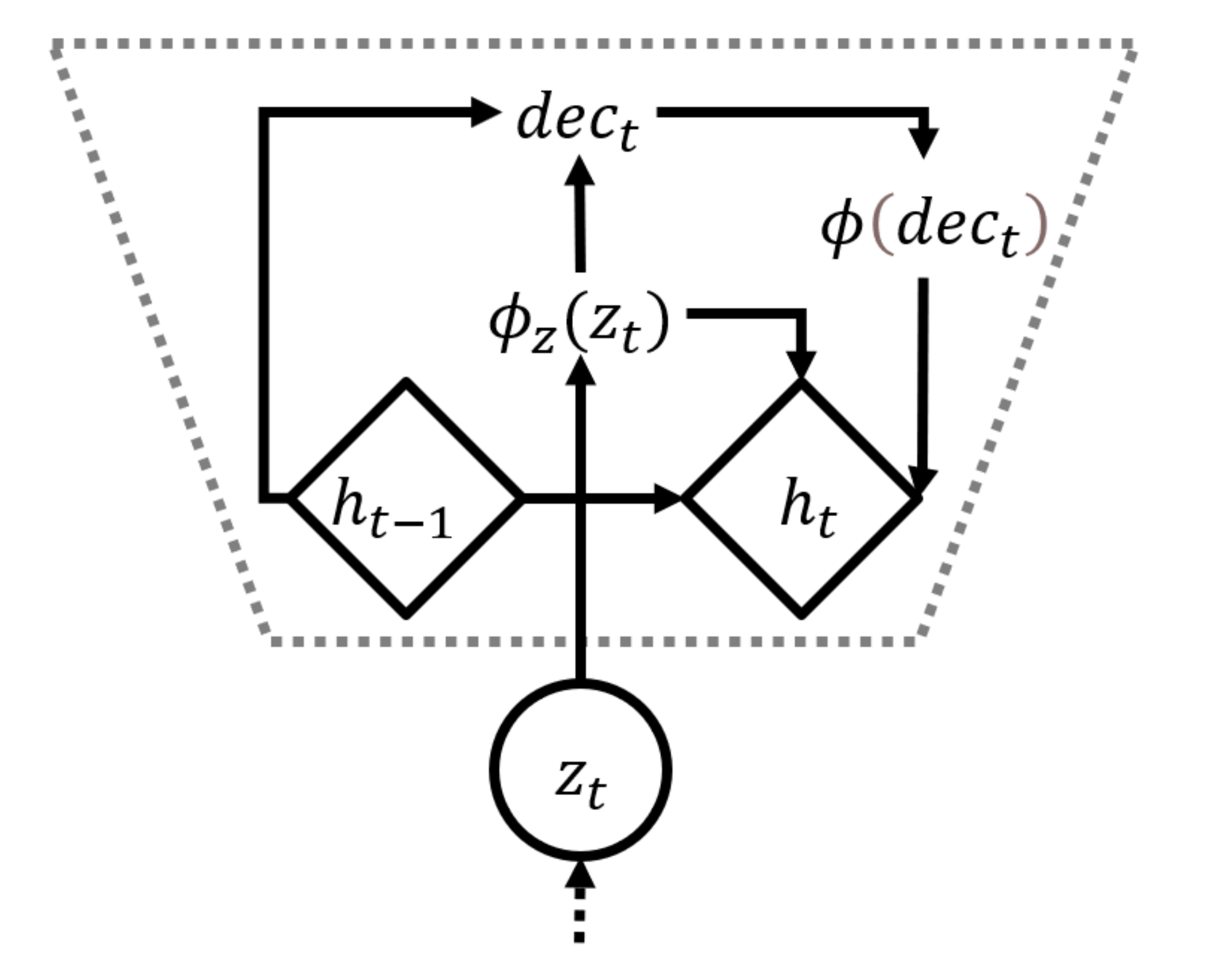}
	\caption{Decoding structure of the VRED during testing. The encoder and decoder can be completely separated via introducing $\phi(dec_t)$. The encoding process is given by Figure \ref{fig:fig1} (b).}
	\label{fig:fig2}
\end{figure}

\subsection{Feature Learning} \label{featlearn}
\par In this work, we aim for an end-to-end DNN audio codec that can work without the use of hand-crafted or biased feature extractors such as LPC (Linear Predictive Coding) coefficients, LPS (Log Power Spectrum) and mel-spectrogram. Instead, we use a simple CNN layer to extract/construct from the input/output of the codec. It has been argued in previous works that the use of feature extractors on top of the model can significantly enhance the performance of the primary model (\cite{chung2015recurrent}). The feature extraction can be understood as a simple pre-processing step, in where the layer aims to capture prominent features of the signal by the use of filters/kernels. It is worth mentioning that maxpooling cannot be used for the compression task, as it downsamples the data and thus harms the reconstruction. The parameters and implementation details are specified in the next section.

\subsection{VRED training}
\par Like the VRNN, the VRED contains a VAE conditioned on the state variable $h_{t-1}$ of an RNN during the generation step (decoder). However, the prior on the latent variable is discrete and follows a Bernoulli distribution: 
\begin{equation} \label{reparam}
    z_t \sim Bern(p_t), \;\;\;\; \mathrm{where} \;\; p_t=\varphi_{prior}(h_{t-1}).
\end{equation}
Then, the generating distribution will not only be conditioned on $z_t$ but also on $h_{t-1}$ such that:
\begin{equation}
    x_t|z_t \sim \mathcal{N}\bigg(p_{x,t}, diag(\sigma_{x,t}^{2})\bigg),
\end{equation}
where $p_{x,t}=dec_t(\phi_z(z_t),h_{t-1})$ and $\sigma_{x,t}^{2} = p_{x,t}(1-p_{x,t})$, $p_{x,t}$ denotes the parameters of the generating distribution, $\varphi_{prior}$ and $dec_t$ can be any highly flexible function such as neural networks. The RNN updates the hidden state using the recurrent equation:
\begin{equation}
    h_t = f_{\theta}(\phi(dec_t),\phi_z(z_t), h_{t-1}),
\end{equation}
where $f$ is the original transition function from Eq. \ref{eq11}. $\phi_z$ and $\phi_{x}$ are NNs that extract features from $z_t$ and $x_t$, respectively. The introduction of $\phi(dec_t)$ allows to decouple the encoder from the decoder, as there is no link between the encoding step and the latent variable during inference (Figure \ref{fig:fig2}). 
\par During inference, the approximate posterior will be a function of $x_t$ and $h_{t-1}$ following the equation: 
\begin{equation}
    z_t|x_t \sim Bern(p_{z,t}), \;\;\;\; \mathrm{where} \;\; p_{z,t}=enc_t(\phi_x(x_t), h_{t-1}),
\end{equation}
where $p_{z,t}$ denotes the parameter of the approximate posterior. We note that the encoding of the approximate posterior and the decoding of generation are tied through the RNN hidden state $h_{t-1}$.
The objective function remains the same timestep-wise variational lower bound as in Eq. \ref{elbo2}, but in this case, the Kullback-Lieber divergence is given by:
\begin{equation} \label{klber}
    KL(q||p)_{Ber} = q \log\frac{q}{p} + (1-q) \log(\frac{1-q}{1-p}),
\end{equation}
where $q$ and $p$ follow a Bernoulli distribution. 
\par To facilitate the comparison, the notations in the equations of this section are similar to \cite{chung2015recurrent}.

\subsubsection{Reparameterization trick}
\par The main issue with using a discretized prior is that backpropagation is hard to flow through the discretized nodes. To solve this, we use the same parametrization trick presented in \cite{raiko2014techniques}. 
\par The main idea is that given a sample $t\sim Bern(p)$ as in Eq. \ref{reparam}, we define a quantity $c=t(1-p) - (1-t)p$ and detach it from the computation graph. Instead of backpropagating through $t$, we use $p+c$ equivalently and avoid the sampling block this way.

\subsection{Fine-tuning}
\par After training the VRED, we de-freeze the CNN codec layer and fine-tune the whole model with a single objective function which is given by Eq. \ref{elbo2} and Eq. \ref{klber},
\begin{equation}
 \mathbb{E}_{q(\mathbf{z}_{\leq T}|\mathbf{x}_{\leq T})} \bigg[ \sum_{t=1}^{T}(-q(z_t|x_{\leq t}, z_{<t}) \log\frac{q(z_t|x_{\leq t}, z_{<t})}{p(z_t|x_{<t}, z_{<t})} + (1-q(z_t|x_{\leq t}, z_{<t})) \log(\frac{1-q(z_t|x_{\leq t}, z_{<t})}{1-p(z_t|x_{<t}, z_{<t})}) + \log p(\mathbf{x}_t|\mathbf{z}_{\leq t}, \mathbf{x}_{<t}))\bigg].
\end{equation}
\par
\begin{figure}[ht]
	\centering
	\includegraphics[width=0.45\linewidth]{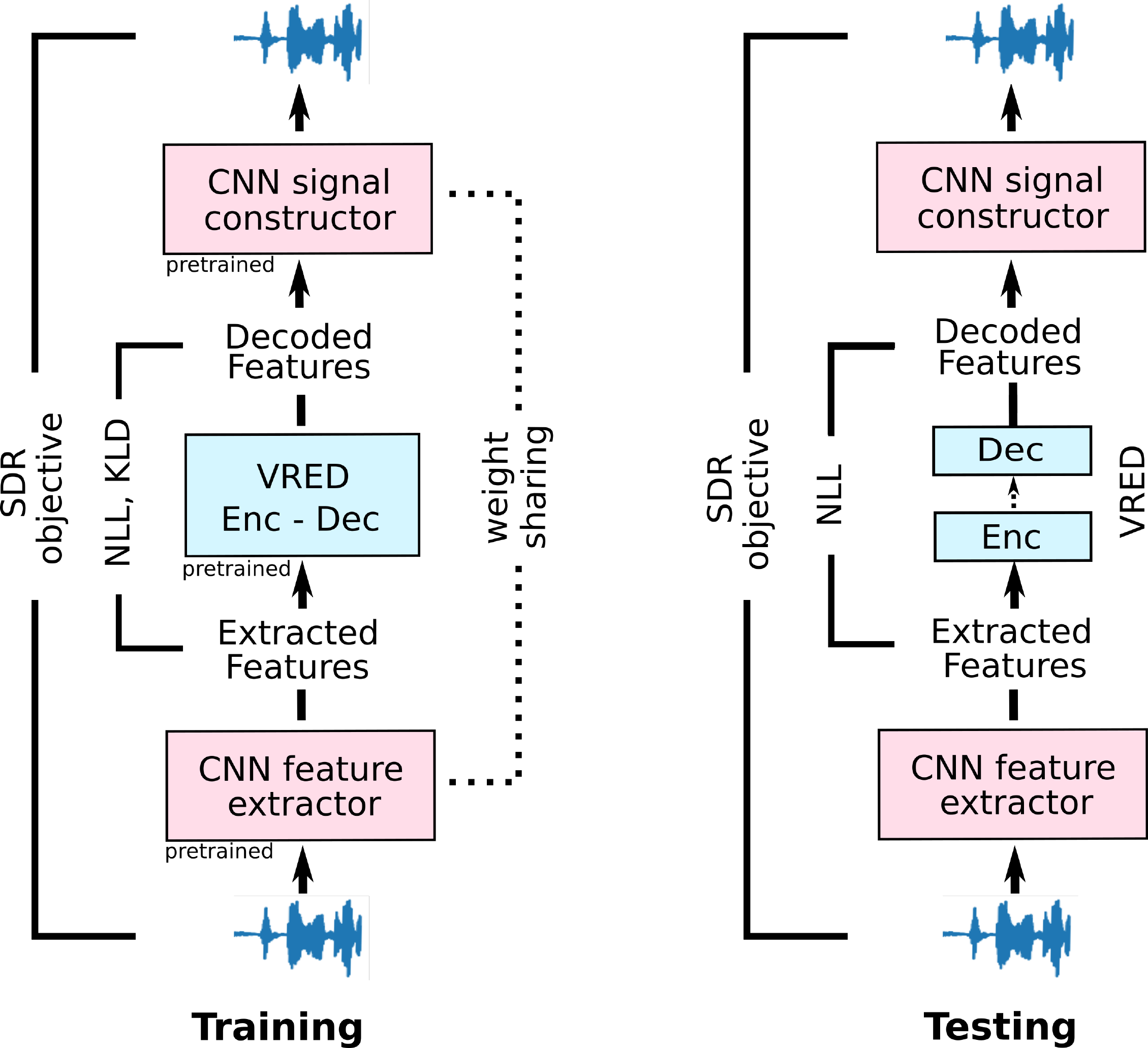}
	\caption{End-to-end model used for training and testing. Note that during testing the encoder and decoder are separable.}
	\label{fig:fig3}
\end{figure}

 The end-to-end architecture is shown in Figure \ref{fig:fig3}.

\section{Experiments}
\par To train the VRED model, we used the BBC sound effects dataset \footnote{\textit{BBC sound effects library}, \url{http://www.sound-ideas.com/sound-effects/bbc-sound-effects.html./}, 2015} sampled at 44.1kHz, which consists of 15,000 audio files of different lengths (cars passing, plane engines, metal objects falling). For consistent processing and batching, we split the audio files into chunks of 0.02s. For the testing dataset, we did not use a partition of the training dataset, as we wanted to try the compression of more complex audio files. We used 15 audio files between 8s and 17s, including music, speech over noise, single instruments, and others \footnote{\textit{Evaluation Guidelines for Unified Speech and Audio Proposals}, ISO/IEC SC29 WG11 N9638, MPEG, Jan. 2008.}. 
\par For both sets, the format of the audio files is \textit{``.wav"}, and the sampling rate is 44.1 kHz. The signals are converted from stereo (2 channels) to mono. 
\par To determine the quality of the reconstructed signal, we used the Signal-to-Distortion Ratio score (\cite{nakajima2018monaural}). The score is in decibels (dB), and reflects how similar the estimated signal $\hat{s}$ is to the clean signal $s_{target}$. It is defined as 
\begin{equation}
    SDR = 10 \log_{10} \frac{||s_{target}||^2}{||\hat{s}-s_{target}||^2}.
\end{equation}
The higher the SDR score means the higher the recovery rate between the estimated and target signals). For example, an SDR  60dB is considered a nearly perfect reconstruction, and a reconstruction with an SDR of around 30dB is considered successful.
\par For the feature extraction of the raw signals, we used a one-dimensional convolution layer. As the signal constructor we used the deconvolution version of the CNN feature extractor. We experimented with several number of kernels and kernel dimensions, which will be detailed in next section. To learn the parameters of the prior, approximate posterior and generative models we used feedforward networks. The encoder and decoder of the VRED consisted in three feedforward layers with hidden dimension 128, which was also the dimension of latent variable $\mathbf{z}$. We applied recurrences via LSTM. The Bernoulli prior consisted in two linear layers. For fine-tuning we used Adam optimization with learning rate re-scheduling.

\section{Results}
\subsection{Feature extractor}
\par The experiment results of the CNN features without VRED are shown in Table \ref{tab:table1}. Several combinations of stride and kernel sizes were tried. We also varied the number of filters. Some results achieved an SDR of almost 48 for our test dataset, although the training SDR was not very high in comparison. What we want is a configuration that would have a high SDR for both the training and testing sets while using a low kernel number to achieve a lower compression rate.

\begin{table}

	\centering
\begin{tabular}{ccccc}
\hline
\textbf{Stride}     & \textbf{Kernel Size} & \textbf{N Kernels} & \textbf{SDR train} & \textbf{SDR test} \\ \hline
\multirow{2}{*}{44} & \multirow{2}{*}{88}  & 32                 & \textbf{23.49}     & \textbf{39.09}    \\ \cline{3-5} 
                    &                      & 64                 & \textbf{35.10}     & \textbf{41.45}    \\ \hline
\multirow{2}{*}{22} & \multirow{2}{*}{44}  & 32                 & 29.07              & 38.14             \\ \cline{3-5} 
                    &                      & 64                 & 31.08              & 43.44             \\ \hline
\multirow{2}{*}{4}  & \multirow{2}{*}{88}  & 32                 & 18.06              & 32.54             \\ \cline{3-5} 
                    &                      & 64                 & 16.97              & 32.19             \\ \hline
\multirow{2}{*}{10} & \multirow{2}{*}{21}  & 256                & 15.63              & 42.33             \\ \cline{3-5} 
                    &                      & 128                & 21.09              & 47.21             \\ \hline
\multirow{2}{*}{44} & \multirow{2}{*}{100} & 32                 & 25.30              & 42.33             \\ \cline{3-5} 
                    &                      & 64                 & 29.37              & 38.37             \\ \hline
\multirow{2}{*}{44} & \multirow{2}{*}{80}  & 32                 & 22.14              & 37.67             \\ \cline{3-5} 
                    &                      & 64                 & 27.4               & 37.50             \\ \hline
\end{tabular}
	\centering
	\vspace{1em}
	\caption{SDR for training and testing datasets when training only the CNN feature extraction of VRED (Section \ref{featlearn}) for 500 epochs. The results come from different configurations of the CNN structure.}
	\label{tab:table1}
\end{table}

\subsection{VRED activations}
\par
\begin{figure}[ht]
	\centering
	\includegraphics[width=0.55\linewidth]{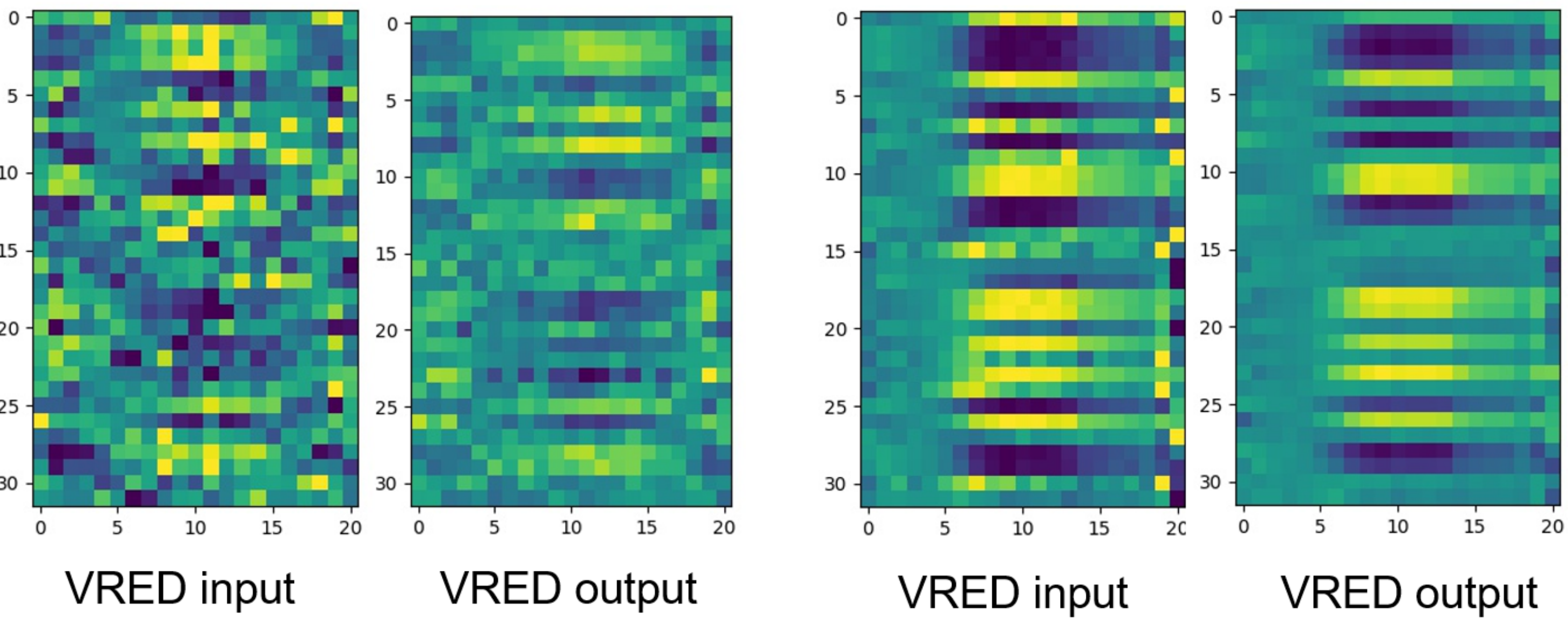}
	\caption{Examples of VRED input and output for different encoded signals, after 3000 epochs.}
	\label{fig:fig4}
\end{figure}

 The VRED is trained using the pre-trained feature extractor with the best train/testing SDR value configuration. If we focus on the VRED training part only (Section 3.2), we analyze the input to the model and output of the RNN decoder.  From Figure \ref{fig:fig4}, we observe how the location of bright and dark pixels representing the extracted features values are somewhat correctly located. Even though the reconstruction may not seem perceptually different, the SDR score is directly proportional to the difference in the target and reconstructed features. 
\subsection{Fine-tuned VRED results}
We fine-tuned VRED with a CNN feature extractor of 32 kernels with size 88 and a stride of 44. The dimension of the latent space was set to 128. The total compression rate of the audio signals resulted in 
\begin{equation}
    CR = CR_{CNN} \times CR_{VRED} = \frac{32}{44} \times \frac{128}{(32\times 32)} = \frac{1}{11},
\end{equation}
where $CR_{CNN}$ is the compression of the feature extractor and $CR_{VRED}$ corresponds to the compression achieved by the VRED.
\par
\begin{figure}[ht]
	\centering
	\includegraphics[width=0.70\linewidth]{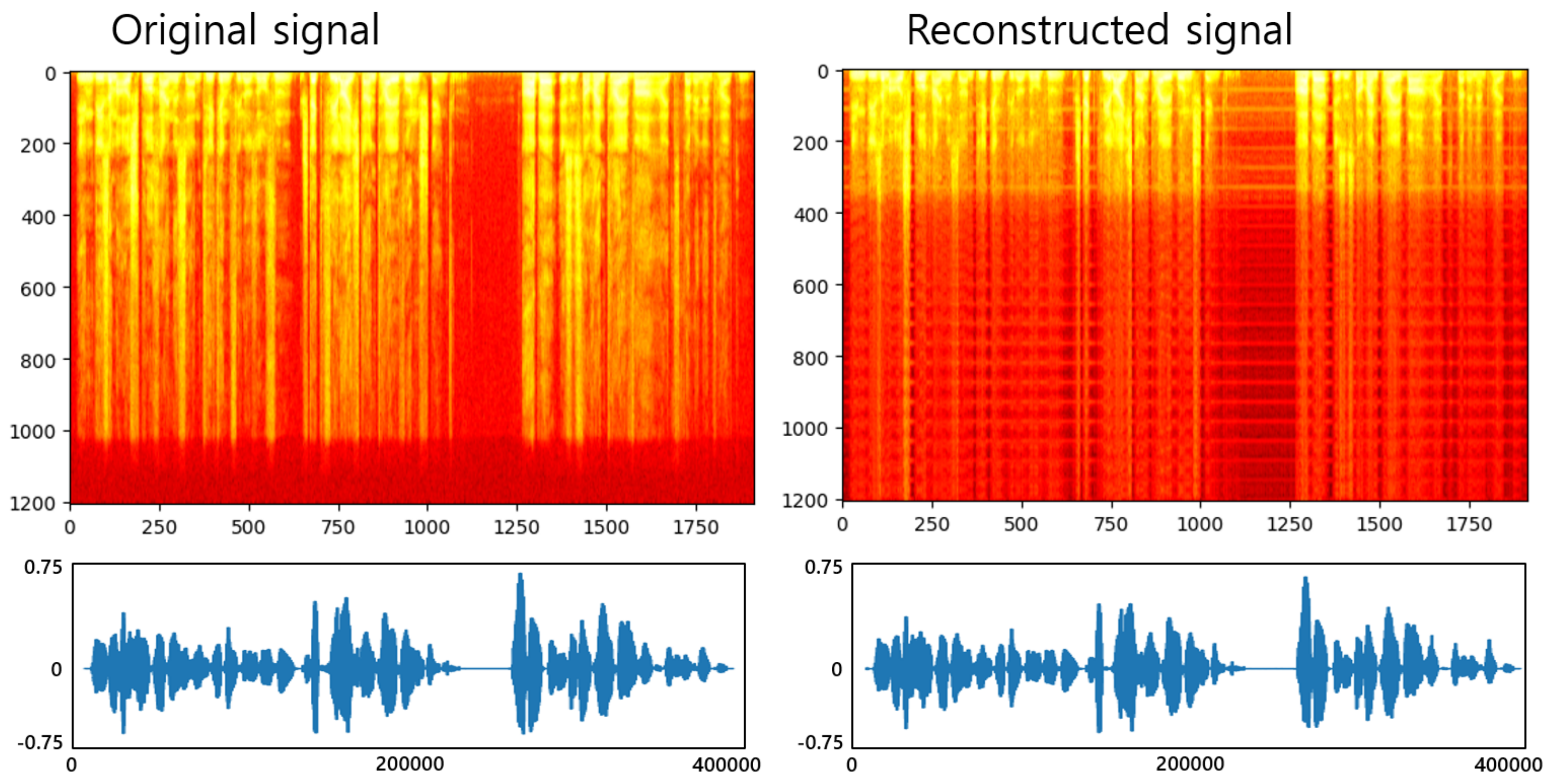}
	\caption{Example of a test audio reconstruction after compression. We present the spectrogram of the signals (top) and their amplitudes through time (bottom). The reconstruction signal is noisier than the original signal. Furthermore, the non-smooth bands in the reconstruction spectrogram suggest that the model does not capture the original frequency spectrum in high detail.}
	\label{fig:fig5}
\end{figure}

 The SDR after the fine-tuning reached 20.53 dB. An example of the reconstruction is shown in Figure \ref{fig:fig5}.

\section{Conclusion}
In the present work, we showed a novel, completely separable end-to-end model using deterministic latent variables. Our model is simpler and more straightforward to train than previous models. Though the obtained results are not commercially acceptable, the proposed approach opens a new door for researching audio compression using unified deep learning models with end-to-end learning. 

\section{Acknowledgments}
This work was supported by Electronics and Telecommunications Research Institute (ETRI) grant funded by the Korean government. [21ZH1200, The research of the basic media・contents technologies]

\bibliographystyle{unsrtnat}
\bibliography{VREDtex}  






\end{document}